\documentclass[11pt,a4paper]{article}
\usepackage[hyperref]{acl2021}
\usepackage{times}
\usepackage{latexsym}
\usepackage{graphicx}
\usepackage{booktabs}
\usepackage{mdframed}
\usepackage{tikz}

\definecolor{orange}{HTML}{f97306}

\usepackage{microtype}

\aclfinalcopy % Uncomment this line for the final submission
\def\aclpaperid{3789} %  Enter the acl Paper ID here

\title{Investigating Memorization of Conspiracy Theories in Text Generation\\ 
{\small \textcolor{orange}{Note: This paper contains examples of potentially offensive conspiracy theory text.}}}

\author{Sharon Levy, Michael Saxon, William Yang Wang \\
  University of California, Santa Barbara \\
  \texttt{sharonlevy@cs.ucsb.edu, saxon@ucsb.edu, william@cs.ucsb.edu} \\}

\date{}

\begin{document}
\maketitle
\begin{abstract}
The adoption of natural language generation (NLG) models can leave individuals vulnerable to the generation of harmful information memorized by the models, such as conspiracy theories. While previous studies examine conspiracy theories in the context of social media, they have not evaluated their presence in the new space of generative language models. In this work, we investigate the capability of language models to generate conspiracy theory text. Specifically, we aim to answer: can we test pretrained generative language models for the memorization and elicitation of conspiracy theories without access to the model's training data? We highlight the difficulties of this task and discuss it in the context of memorization, generalization, and hallucination. Utilizing a new dataset consisting of conspiracy theory topics and machine-generated conspiracy theories helps us discover that many conspiracy theories are deeply rooted in the pretrained language models. Our experiments demonstrate a relationship between model parameters such as size and temperature and their propensity to generate conspiracy theory text. These results indicate the need for a more thorough review of NLG applications before release and an in-depth discussion of the drawbacks of memorization in generative language models.
\end{abstract}

\section{Introduction}
Recent advances in natural language processing technologies have opened a new space for individuals to digest information. One of these rapidly developing technologies is neural natural language generation. These models, made up of millions, or even billions~\cite{NEURIPS2020_1457c0d6}, of parameters, train on large-scale datasets. While attempts are made to ensure that only ``safe'' data is utilized for training these models, several studies have shown the prevalence of biases produced by these pretrained generation models~\cite{sheng-etal-2019-woman,groenwold-etal-2020-investigating,solaiman2019release}. Of equally alarming concern are the memorization and subsequent generation of factually incorrect data. Conspiracy theories are one particular type of this data that can be especially damaging. 

While it is not new for researchers to learn that a model may memorize data~\cite{radhakrishnan2019memorization}, we argue that the growing usage of machine learning models in society warrants targeted investigation to deter potential harms from problematic data. In this paper, we address the upsides and pitfalls of memorization in generative language models and its relationship with conspiracy theories. We further describe the difficulty of detecting this memorization for the categories of memorization, generalization, and hallucination. Previous studies investigating memorization of text generation models have done so with access to the model's training data \cite{carlini2019secret,carlini2020extracting}. As models are not always published with their training datasets, we set out to examine the difficult task of eliciting memorized conspiracy theories from a pretrained NLG model through various model settings \textbf{without access to the model's training data}. 

 We focus our study on the pre-trained GPT-2 language model~\cite{radford2019language}. 
We investigate this model's propensity to generate conspiratorial text, analyze relationships between model settings and conspiracy theory generation, and determine how these settings affect the linguistic aspect of generations. To do so, we create a new conspiracy theory dataset consisting of conspiracy theory topics and machine-generated conspiracy theories.

Our contributions include:
\begin{itemize}
  \item[$\bullet$]  We propose the topic of conspiracy theory memorization in pretrained generative language models and outline the harms and benefits of different types of generations in these models. 
  \item[$\bullet$] We analyze pretrained language models for the inclusion of conspiracy theories without access to the model's training data.
  \item[$\bullet$] We evaluate the linguistic differences for generated conspiracy theories across different model settings.
  \item[$\bullet$] We create a new dataset consisting of conspiracy theory topics from Wikipedia and machine-generated conspiracy theory statements from GPT-2.
\end{itemize}

\section{Spread of Conspiracy Theories}
\subsection{Dangers of conspiracy theories}
A conspiracy theory is the belief, contrary to a more probable explanation, that the true account for an event or situation is concealed from the public~\cite{goertzel1994belief}. 
A variety of conspiracy theories ranging from the science-related moon landing hoax~\cite{bizony2009all} to the racist and pernicious Holocaust denialism\footnote{http://auschwitz.org/en/history/holocaust-denial/} are widely known throughout the world.
However, even as existing conspiracy theories continue circulating, new conspiracy theories are consistently spreading. This is especially concerning given that half of Americans believe at least one conspiracy theory~\cite{oliver2014conspiracy}. 

Widespread belief in conspiracy theories can be highly detrimental to society, driving prejudice \cite{douglas2019understanding}, inciting violence\footnote{https://www.theguardian.com/us-news/2019/aug/01/conspiracy-theories-fbi-qanon-extremism}, and reducing science acceptance~\cite{VANDERLINDEN2015171,lewandowsky2013nasa}. Science denial has real-world consequences, such as resistance to measures for the reduction of carbon footprints~\cite{doi:10.1177/0096340215571908} and outbreaks of preventable illnesses due to reduced vaccination rates~\cite{goertzel2010conspiracy}. Further effects of conspiracy theory exposure can reach the political space and reduce citizens' likelihood of voting in elections due to feelings of powerlessness towards the government~\cite{jolley2014social}. 

At the time of writing, the COVID-19 pandemic is at its worst. Though COVID-19 vaccines have received approval and started distribution, new conspiracy theories surrounding the COVID-19 vaccine may hinder society in its road to recovery. Discussions of a link between vaccinations and autism have been circulating for years~\cite{jolley2014effects,kata2010postmodern}. However, with the extreme interest throughout the world surrounding the COVID-19 pandemic, new vaccination rumors are arising, such as the vaccine causing DNA alteration and claims of the pandemic acting as a cover plan to implant trackable microchips\footnote{https://www.bbc.com/news/54893437}. 
The belief in these theories can prevent herd immunity through the lack of vaccinations\footnote{https://www.economist.com/graphic-detail/2020/08/29/conspiracy-theories-about-covid-19-vaccines-may-prevent-herd-immunity}~\footnote{https://www.who.int/news-room/q-a-detail/herd-immunity-lockdowns-and-covid-19}.

\subsection{NLG spreading conspiracy theories}
As NLG models are being utilized for various tasks such as chatbots and recommendations systems~\cite{gatt2018survey}, cases arise in which these conspiracy theories and other biases can propagate unintentionally~\cite{bender2021dangers}. We present one such scenario in which an NLG model has memorized some conspiracy theories and is being used for story generation~\cite{fan-etal-2018-hierarchical}. An unaware individual may utilize this application and, given a prompt about the Holocaust, may receive a generated story discussing Holocaust denial. The user, now having been exposed to a new conspiracy theory, may choose to ignore this generated text at this stage. However, a potential negative outcome is that the user may become interested in this story and search the statements online out of curiosity. This can lead the user down the ``rabbit hole'' of conspiracy theories online~\cite{o2015down} and alter their original assumptions towards believing this conspiracy theory.  

\subsection{Why are conspiracy theories difficult to detect?}\label{sec:difficult}
Recent years have seen the emergence of several new tasks addressing fairness and safety within natural language processing in topics such as gender bias and hate speech detection. Although detection and mitigation of other biases and harmful content have been thoroughly studied, that pertaining to conspiracy theories is increasingly difficult due to its inconsistent linguistic nature. 

Many existing tasks can utilize specific keyword lists such as Hatebase\footnote{https://hatebase.org/} for detection in addition to current techniques~\cite{sun2019mitigating}. However, conspiracy theory detection is an increasingly complex problem and cannot be approached in the same way as the previous topics. Conspiracy theories have no unified vocabulary or keyword list that can differentiate them from standard text. Previous studies of conspiracy theories have exhibited their tendency to lean towards issues of hierarchy and abuses of power~\cite{klein2019pathways}. We argue this is not specific enough to define features for their detection. Often, specific keywords and tropes become typical of conspiracy theories regarding a specific topic, such as 9/11 and ``false-flag''~\cite{knight2008outrageous}. However, as the number of topics surrounding conspiracy theories grows, it becomes infeasible to create and maintain these topic-specific vocabularies. To add to this difficulty, while humans can typically detect other types of biases, they cannot easily distinguish conspiracy theories from truthful text by merely reading the statement. Doing so typically requires knowledge of the topic itself or a more in-depth look into the theory narrative through network analysis\footnote{https://theconversation.com/an-ai-tool-can-distinguish-between-a-conspiracy-theory-and-a-true-conspiracy-it-comes-down-to-how-easily-the-story-falls-apart-146282}. To this end, the best way to stop the spread of conspiracy theories is not in late-stage detection but early intervention.

\subsection{How can NLG models be misused?}
While the generation of conspiracy theories may be an accidental outcome by NLG models, the possibility still exists that adversaries will intentionally utilize these language models to spread these theories and cause harm. In one such case, propagandists may utilize NLG models to reduce their workload when spreading influence~\cite{mcguffie2020radicalization}. By merely providing topic-specific prompts, they can utilize these models to easily and efficiently produce a variety of conspiratorial text for online communities regarding the topics. As a result, these communities will appear to be larger than their actual size and provide the appearance that belief in the issue is high. This may provide real-life members with a sense of belonging and subsequently reinforce belief in the theories or even recruit new members~\cite{douglas2017psychology}.

\section{Memorization vs. Generalization vs. Hallucination}
The memorization of data in the context of machine learning models has been highlighted in research for many years now. Related work has researched the types of information models memorize~\cite{NEURIPS2020_1e14bfe2}, how to increase generalization~\cite{chatterjee2018learning}, and the ability to extract information from these models~\cite{carlini2020extracting}.  While memorization is typically discussed in the space of memorization vs. generalization, we believe this can be broken down even further. In the context of conspiracy theories, we establish three types of generations:
\begin{itemize}
  \item[$\bullet$] Memorized: generated conspiracy theories with exact matches existing within the training data.
  \item[$\bullet$] Generalized: generations that do not have exact matches in the data but produce text that follows the same ideas as those in the training data.
  \item[$\bullet$] Hallucinated: generations about topics that are neither factually correct nor follow any of the existing conspiracy theories surrounding the topic.
\end{itemize}
Studies on memorization tend to focus on either memorization vs. generalization or memorization vs. hallucination~\cite{nie-etal-2019-simple}. In the latter case, it is easy to see how the term ``memorization'' can apply to the first two categories. Ideally, in an NLG model, we would hope for generations to be generalized since direct memorization can have the downsides of generating sensitive information~\cite{carlini2019secret}. There are also cases when hallucinations are ideal, such as in the realm of creative story-telling. Should we be able to distinguish among these categories, we could gain deeper insight into what and how these models learn during training. However, we acknowledge that classifying generations based on these categories is a difficult problem and believe this should be a task for future research in memorization.

Our focus in this paper is to evaluate 1) whether a model has memorized conspiracy theories during training and 2) the propensity for the model to generate this information among different model settings (as opposed to generating other memorized or hallucinated information about a topic). Evaluating memorization within a model can be done in two settings: with training data as a reference or without training data. Previous studies have evaluated memorization within machine learning models by utilizing the model's training dataset. However, the reality is that many models nowadays are not published alongside their training data~\cite{45189}. In this case, the evaluation becomes increasingly difficult, as there is nothing to match a model's output to. In order to simulate a real-world environment, we analyze the second setting of investigating memorization without access to training data and instead treat the model as a black box when evaluating its outputs. Due to the difficulty of distinguishing among the three categories of memorization, generalization, and hallucination, we follow previous work and refer to both memorized and generalized generations as memorized samples for the rest of the paper.

\section{When is memorization a good thing?}
While we focus most of this paper on the downsides of memorization in natural language generation models, it is still important to address the benefits. There are several situations in which memorized information may be utilized, such as in dialogue generation~\cite{gu-etal-2016-incorporating}. When used in the chatbot setting, a model may be asked questions on real-world knowledge. Assuming the model has learned correct factual information, this memorization can prove useful. Furthermore, conspiracy theories are a part of language and culture. It is not inherently bad that a model is aware of the existence or concept of conspiracy theories, particularly in cases where models may be deployed as an intervention in response to human-written conspiratorial text. This only becomes harmful when the model cannot recognize text as a conspiracy theory and generates text from the viewpoint of the conspiracy being true. Though memorization may aid in the described cases, the downside of the learned conspiracy theories (as factual statements) and other information such as societal biases can outweigh these benefits.

\begin{table}[t]
\centering
\begin{tabular}{p{1.6cm}|p{5.4cm}}
\toprule
&  \multicolumn{1}{c}{Conspiracy Theory} \\
\hline
Wikipedia &  The \textbf{Holocaust} is a lie, and the Jews are not the victims of the Nazis.  \\
\hline
GPT-2 &   
The US government is secretly running a secret program to create a super-soldier that can kill and escape from any prison. \\
\bottomrule
 \end{tabular}
\caption{Samples from the Wikipedia dataset consisting of Wikipedia topics and General dataset of GPT-2 generated conspiracy theories without topic prompts. The Wikipedia topic is highlighted in bold and is used as a topic-prompt for text generation in GPT-2.
}\label{tab:example}
\end{table}

\section{Data Collection}
While conspiracy theory data may appear in misinformation datasets labeled as ``Fake News'' with other misinformation types, there are few existing datasets with conspiracy theory labeled text. Previous conspiracy theory studies contain datasets that are either small in size~\cite{oliver2014conspiracy}, contain non-English data~\cite{bessi2015science}, or pertain to events occurring after the release of GPT-2~\cite{ahmed2020covid,uscinski2020people}. Therefore, we create a dataset exclusively dedicated to conspiracy theories~\footnote{https://github.com/sharonlevy/Conspiracy-Theory-Memorization}. We obtain our data for our analysis from two different sources: Wikipedia and GPT-2. We show samples from each of our datasets in Table \ref{tab:example}.
\subsection{Wikipedia}
We first aim to create a set of conspiracy theory topics. To gather this data, we utilize Wikipedia's category page feature. Each item listed in a category page is linked to a corresponding Wikipedia page. We obtain the page headers in the conspiracy theory category page and the following page headers in the Wikipedia conspiracy theory category tree. This process allows us to extract 257 Wikipedia pages regarding conspiracy topics. We further refine this dataset of conspiracy topics through the use of Amazon Mechanical Turk~\cite{buhrmester2016amazon}. Ten workers are assigned to each Wikipedia conspiracy topic, and each worker is asked whether they have heard of a conspiracy theory related to the topic. We remove any topic from our dataset with fewer than six votes to focus our study on the well-known conspiracy theory topics that a model would be more likely to be prompted with. Our final dataset consists of the following seventeen conspiracy theory topics: Death of Marilyn Monroe, Men in black, Sandy Hook school shooting, UFO's, Satanic ritual abuse, Climate change, Area 51, 9/11, Vast right-wing conspiracy, Global warming, Shadow government, Holocaust, Flat Earth, Illuminati, Pearl Harbor, Moon landing, and John F. Kennedy assassination. We refer to this as the Wikipedia dataset for the remainder of the paper.

\subsection{GPT-2}
We create a second dataset consisting of machine-generated conspiracy theories. To do this, we elicit the conspiracy theories directly from GPT-2 Large with the HuggingFace transformers library~\cite{wolf-etal-2020-transformers}. We prompt GPT-2 with ``The conspiracy theory is that'' at varying temperature levels (0.4, 0.7, 1). We obtain 5000 theories at each temperature level and post-process the text by removing the original prompt and keeping only the first sentence. For the remainder of the paper, we refer to this dataset as the General dataset.

\section{Generation of Conspiracy Theories}\label{sec:context_gen}

An intriguing question in the scope of conspiracy theory generation is: what can trigger a language model to generate conspiracy theories? We begin by investigating the effects of model parameters and decoding strategies on the generation of conspiracies when prompted with a topic. Of these, we study model temperature and model size. 

We use our Wikipedia dataset to create a generic prompt as input to GPT-2, such as ``The Holocaust is''.  In order to remove any trigger such as “Flat Earth is”, we modify some of our topic titles during prompt creation to make a more neutral prompt. In the case of “Flat Earth”, our prompt is “The Earth is”, so that the model is not intentionally triggered to produce Flat Earth conspiracy text. We perform this action for the rest of our topics as well. For each prompt, we employ the model to create twenty generations with a token length of fifty. 

When evaluating the generated text, we evaluate whether or not the text affirms the conspiracy theory. In this sense, we count ``The Earth is flat'' as affirming the conspiracy theory and “The Earth is flat is a conspiracy theory” as not affirming the theory. As such, we evaluate whether the model presents the theory as factual belief as opposed to whether it has knowledge of the theory. 

To determine whether or not the generation affirms a known conspiracy theory, we utilize Amazon Mechanical Turk. Worker pay and instructions are detailed in Appendix \ref{sec:appendix}. We provide each worker with a reference passage describing known conspiracy theories for each topic and ask whether or not the generation affirms or aligns with the reference text. We make sure to state that the reference text contains several conspiracy theories about the topic at the top of each HIT. In this case, if a worker is exposed to new text, they are clearly informed that the text is a conspiracy theory. Should a worker encounter these theories in the future, they may even benefit from the task since they are now armed with the knowledge that these statements are in fact conspiracy theories. Seven workers are assigned to each generated sequence. If the text is voted as a conspiracy, it receives a point; otherwise, it is subtracted a point. We then retrieve those generations with two or more points (indicating a general consensus) and manually evaluate this subset of generations for another round of verification.

\subsection{Temperature}

\begin{figure}[t]
  \centering
  \includegraphics[width=\linewidth]{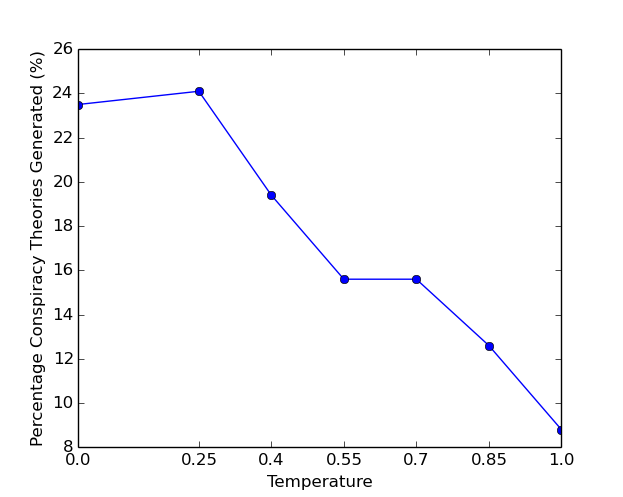}
  \caption{Percentage of conspiracy theories generated by GPT-2 Large at varying temperatures when prompted on 17 different conspiracy theory topics. Each topic is used to generate 20 sequences for a total of 340 generations.}\label{fig:temp}
\end{figure}

 We first evaluate GPT-2 Large at temperature settings ranging from 0.25 to 1 with sampling, where 1 is the default setting for the model, and with greedy decoding on the Wikipedia dataset prompts. This decoding strategy changes the model's probability distribution for predicting the next word in the sequence. A lower temperature will increase the likelihood of high probability words and decrease the likelihood of low probability words. At each temperature level, we compute the percentage of generated text marked as conspiracy theories out of the total number of generations. We share our results in Figure \ref{fig:temp}. 

It can be seen that as the temperature decreases, the model follows a general trend of generating more conspiracy theories. There is an exception when temperature $\rightarrow$ 0, which translates to simple greedy decoding. In this case, the proportion of conspiracy theories decreases slightly, indicating that while the model may memorize some theories, other information for specific topics is also memorized and have a higher likelihood of being generated. However, the general result curve shows that existing conspiracy theories are deeply rooted in the model during training for many topics. Given these findings, we believe it is best to add randomization to the decoding procedure, at the risk of quality and coherency, instead of greedy search in order to minimize the risk of generating deeply memorized conspiracy theories.  

Decreasing the model's temperature allows us to evaluate which topics this deep memorization may be true for, as not every conspiracy topic may be ingrained in the model. We assess which topics the model increases its number of conspiracy theory generations for at a lower temperature. When parsing the previous results for each topic across the different temperature settings, we find this increase in conspiracy theory generations and, therefore, the prominent memorization of conspiracy theories for the topics of UFO's, 9/11, Holocaust, Flat Earth, Illuminati, and Moon landing.

\begin{figure}[t]
  \centering
  \includegraphics[width=0.96\linewidth]{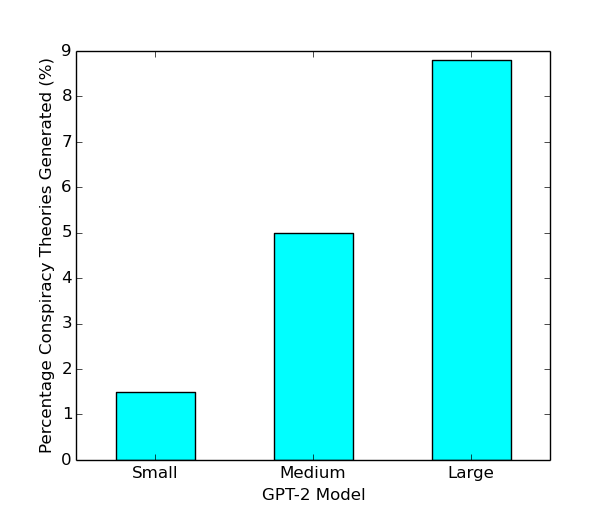}
  \caption{Percentage of conspiracy theories generated by GPT-2 models of size small, medium, and large when prompted on 17 different conspiracy theory topics. Each topic is used to generate 20 sequences for a total of 340 generations.}\label{fig:size}
\end{figure}

\subsection{Model size}
Next, we aim to test a language model's size for its capability to memorize and generate conspiracy theories. Again, we utilize the Wikipedia dataset prompts for generations. We prompt three model sizes with our topics: GPT-2 Small (117M parameters), GPT-2 Medium  (345M parameters), and GPT-2 Large (762M parameters). We keep a fixed temperature across the models and set it at the default value of 1. We use the same evaluation technique described above and compute the proportion of generations marked as conspiracy theories out of the total number of generations. These results are shown in Figure \ref{fig:size}. 

While nearly 10\% of GPT-2 Large's generations are classified as conspiracy theories, GPT-2 Medium reduces this number by almost 50\%. The GPT-2 Small model's conspiracy theory generations are substantially lower than this at a little over 1\%. We can deduce that reducing model size vastly lowers a model's capacity to retain and memorize information after training, even if that information is profoundly prominent within the training data. Not only is this beneficial for mitigating the generation of conspiracy theories, but it can also allow the model to generalize better to other information for topic-specific prompts.

\section{Towards Automated Evaluation}
As we have shown that varying temperature and model size can individually lead to further elicitation and memorization of conspiracy theories, we now investigate the effects of varying the two together. In our previous experiments, we utilize Mechanical Turk to identify conspiracy theories among the generated text. However, we understand that human evaluation is not feasible for detecting conspiracy theories on a large scale. Instead, we desire to advance towards a more automated evaluation of memorization. As such, we investigate whether we can define a relationship between the memorization of conspiracy theories and perplexity across the different model parameters. 

Following previous studies on fact-checking \cite{chakrabarty-etal-2018-robust, Wang2010GotYA} and model memorization~\cite{carlini2020extracting}, we evaluate model generations against Google search results. This time, we utilize our General dataset, made up of conspiracy theories generated with the generic prompt ``The conspiracy theory is that''. We query Google with a generated conspiracy theory at each temperature setting and compare this theory to the first page of results. We did not manually use Google search for our generated text and instead created a script to automate this and scrape the text from the first page of results. We provided the minimum amount of information needed for making each search request so that this does not include search history and the more specific location information or cookies.

\begin{figure}[t]
  \centering
  \includegraphics[width=1.0\linewidth]{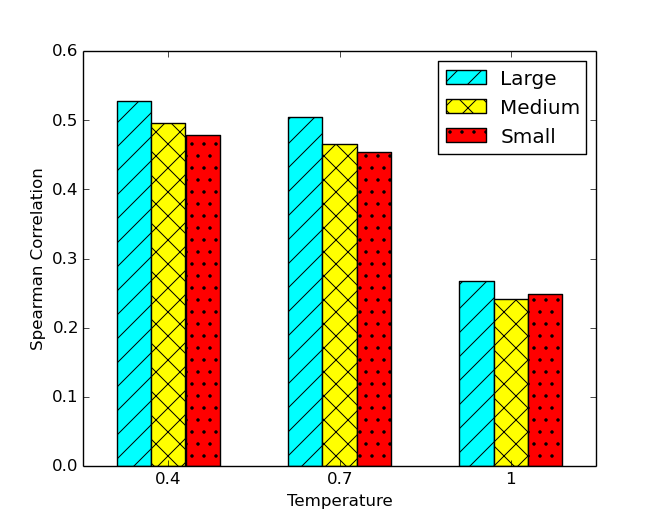}
  \caption{Spearman correlation of model perplexity vs. Google search BLEU score for GPT-2 generated conspiracy theories across varying temperature settings. Each generated theory is evaluated against the first page of Google search results with the BLEU metric.}\label{fig:tempxsize}
\end{figure}

The temperature values of 0.4, 0.7, and 1 are used as lower temperature values start to produce many duplicate generations and lead to small sample sizes for this evaluation. We obtain the text snippet under each search result and evaluate this against the conspiracy theory with the BLEU metric~\cite{papineni-etal-2002-bleu}. The BLEU metric is utilized since many search results do not contain complete sentences and are instead highlighted phrases from the text related to the query and concatenated by ellipses. The perplexity score for a conspiracy theory is then calculated for each model size. The resulting BLEU and perplexity scores are ranked with the highest BLEU and lowest perplexity scores first. We use Spearman's ranking correlation~\cite{hogg2005introduction} to determine the resulting alignment between the two. These results are shown in Figure \ref{fig:tempxsize}. 

We find a strong relationship between a generated conspiracy theory's perplexity and its appearance in Google search results. This correlation becomes much weaker when the temperature is set to 1, indicating that the default setting's increased randomness may produce more hallucinated generations. However, given these results, we believe this can open the door towards the creation of more automated memorization evaluation techniques. Though our samples are generated through GPT-2 Large, we further test this alignment on the small and medium model sizes. We find that the relationship between Google search results and perplexity decreases as model size decreases for the smaller temperature settings, further confirming that model size does affect memorization.

\begin{table}[t]
\centering
\begin{tabular}{l|*{3}{c}|c}
\toprule
&   \multicolumn{3}{c}{Temperature} \\
\hline
Classifier  & 0.4  & 0.7  & {1.0} & p-val   \\
\toprule
dBERT  & -0.974  & -0.942  & -0.887 & 0.110  \\
\hline
VADER  & -0.556  & -0.527  & -0.486 & $<$0.001   \\
\hline
TextBlob  &  -0.112  & -0.033  & 0.017 & $<$0.001  \\
\hline
\hline
Average   & -0.547  & -0.500 & -0.452 \\ 
\bottomrule
 \end{tabular}
\caption{Comparison of average sentiment scores across GPT-2 Large generated conspiracy theories with the DistilBERT (dBERT), VADER, and TextBlob sentiment classifiers along with the Wilcoxon rank-sum p-values for generation pairs of temperature 0.4 and 1. The conspiracy theories are generated at the temperature values of 0.4, 0.7, and 1.0 and sentiment scores range from -1 to 1.
}\label{tab:sentiment}
\end{table}

\section{Linguistic Analysis}
While our previous analysis aims to define a relationship between model parameters and the generation of conspiracy theories, we are also interested in evaluating whether these generations have any interesting linguistic properties. As such, we choose to test the question, are there any linguistic differences among the generated conspiracy theories across different model settings? We proceed by examining two linguistic aspects of our texts: sentiment and diversity. 

\subsection{Sentiment}
When analyzing sentiment, we evaluate our General dataset of generated conspiracy theories at its three temperature levels. We are interested in answering the question: how will the model's temperature affect the sentiment of its generations that are not prompted by real-world stimulus? To proceed, we utilize three sentiment classifiers: DistilBERT~\cite{sanh2019distilbert}, VADER~\cite{gilbert2014vader}, and TextBlob\footnote{https://textblob.readthedocs.io/en/dev/index.html}. For DistilBERT we convert the output range of [0,1] to [-1,1] to match the other two classifier ranges. The average sentiment scores are displayed in Table \ref{tab:sentiment} along with the Wilcoxon rank-sum p-values for each classifier output between temperature settings 0.4 and 1. The results show that decreasing the model's temperature triggers it to generate increasingly negative conspiracy theories. Although we do not achieve similar sentiment scores across the different classifiers, they all exhibit the same downward trend among score and temperature values. Additionally, classifier-temperature value pairs produce negative sentiment scores in all but one case.  This follows previous work indicating that conspiracy theories and one's belief in them are emotional rather than analytical and are linked to negative emotions~\cite{van2018belief}.

\begin{table}[t]
\centering
\begin{tabular}{l|*{3}{c}}
\toprule
&   \multicolumn{3}{c}{Temperature} \\
\hline
Size  & 0.4  & 0.7  & {1.0}   \\
\toprule
Small  & 0.372  & 0.227  & 0.084  \\
\hline
Medium  & 0.397  & 0.231  & 0.094   \\
\hline
Large  &  0.421  & 0.255 & 0.120   \\
\hline
\hline
p-value & $<$0.001 & $<$0.001 & $<$0.001 \\
\bottomrule
 \end{tabular}
\caption{Comparison of average BERTScore values across Wikipedia topic-prompted GPT-2 generations for varying model sizes and temperatures. Generations for each size-temperature pair are evaluated against other generations for their specific topic. Wilcoxon rank-sum p-values for the large-small model pairs at each temperature are listed at the bottom.}\label{tab:diversity}
\end{table}

\subsection{Diversity}
Next, we analyze linguistic diversity across model sizes and model temperature. Utilizing the Wikipedia dataset, we compute the BERTScore~\cite{bert-score} for each generation in reference to the other generations for each topic. This metric is used to measure the variance and contextual diversity across the different model generations for a specific conspiracy topic~\cite{zhu2020understanding}. We do this across temperature values of 0.4, 0.7, and 1 and the different model sizes. These temperature values are utilized as lower temperature values start to produce duplicate generations. The average F1 scores for each setting pair is calculated and shown in Table \ref{tab:diversity} along with the corresponding p-values from a Wilcoxon rank-sum test for the large-small pairs at each temperature.  

We find that as the temperature decreases, the similarity across generations for each topic increases. This is not surprising, as the outputs become less random at lower temperatures, and the model tends to output more memorized information. When comparing the scores among the different model sizes, the largest model contains the largest values, decreasing with the model size. We can infer that an increase in model size leads to more memorization, which allows the model to generate more contextually aligned outputs for specific topics instead of the diverse sets of outputs in smaller model sizes.

\section{Moving Forward}
Throughout this paper, we have discussed the risks and benefits of memorization in NLG models and have focused on the dangers of conspiracy theory generation. As we relayed in Section \ref{sec:difficult}, conspiracy theory detection is a challenging problem due to its fuzzy linguistic vocabulary. We believe it is crucial to intervene earlier to mitigate these risks rather than detect them after the model's generation. While reducing memorization of harmful data in models is still an open problem, we discuss various methods to help accomplish this and encourage future research in the area: 1) preventing detrimental data from being introduced into the training set, 2) ensuring the dataset contains a much larger proportion of factually correct data for conspiracy theory topics than the conspiracy theories themselves, and 3) reducing model size.

The first solution prevents researchers from relying on these models to filter out harmful noise in large-scale datasets. Current models, such as GPT-2, attempt to filter out offensive and sexually explicit content from their datasets during creation\footnote{https://github.com/openai/gpt-2/blob/master/modelcard.md}. We argue that this is not enough, as shown in the results of our analysis above. One way to proceed is to ensure that data is only collected from reliable sources instead of scraping the internet for large amounts of information. However, we also recognize that this is a tedious task and requires intensive scrutiny when collecting data. As such, the downsides to following this method may lead to smaller datasets and models with lower quality generations. In addition, this requires the additional consideration of deciding what data is ``good'' and what data can be harmful. In the space of conspiracy theories, the creation of a database regarding circulated conspiracy theories and debunking them seems like an appropriate direction to go.

While not completely eliminating the possibility of conspiracy theory generation, the second method aims to decrease their likelihood during generation. To accomplish this, researchers can supplement their existing dataset with a second dataset consisting of factually correct samples surrounding conspiracy theory topics. This aims to oversample truthful data for training. While our study is confined to well-known conspiracy theories, the approach we discuss should be performed for all conspiracy-related topics and thus requires the additional task of identifying these subjects.

As our experiments in Section \ref{sec:context_gen} have shown, model temperature and size profoundly affect the memorization and generation of conspiracy theories in NLG models. Since a user may set temperature, this setting cannot help prevent the generation of harmful data. However, modifying model size can. Though recent years have seen an increase in model size due to better performance on downstream tasks and the resulting generation of more coherent text~\cite{solaiman2019release}, it comes at the cost of memorization. Therefore, researchers must strive to find a balance between memorization and fluency. When compromising model size, this mitigation strategy may also be complemented by oversampling factual data as specified above for further intervention.

\section{Conclusion}
In this paper, we highlight the issue of conspiracy theory memorization and generation in pretrained generative language models. We show that the root of the problem stems from the memorization of these theories by NLG models and discuss the dangers that may follow this. This paper further investigates the detection of conspiracy theory memorization in these models in a real-world scenario where one does not have access to the training data. To do so, we create a conspiracy theory dataset consisting of conspiracy theory topics and machine-generated text. Our experiments show that reducing a model's temperature and increasing its size allows us to elicit more conspiracy theories, indicating their memorization without verification against the ground-truth dataset. We hope our findings encourage researchers to take additional steps in testing language models for the generation of harmful content before release. Further, we hope our discussion on memorization can lead to further research in the area and advance the study of conspiracy theories in NLP.

\section*{Acknowledgements}
We would like to thank Amazon AWS Machine Learning Research Award and Amazon Alexa Knowledge for their generous support. This work was also supported in part by the National Science Foundation Graduate Research Fellowship under Grant No. 1650114.

\section*{Ethical Considerations}
In this paper, we explore the topic of conspiracy theories in natural language generation. We acknowledge that in order to build a coherent and robust language model, a large-scale dataset must be used. As such, it is difficult to obtain data of this size that is 100\% free of offensive or harmful samples. However, to improve and further progress research in natural language processing, considerations of disadvantages such as the memorization and subsequent generation of conspiracy theories must be taken into account. 

In the previous sections of the paper, we feature how researchers can evaluate language models for the memorization and generation of conspiracy theory text. While we showcase methods for analyzing the generation of this harmful content, we acknowledge some potential risks: 1) adversaries may adjust their methods for hiding harmful content in language models so that it is not easily extracted or generated through our evaluation methods, and 2) attackers may use the methods discussed and leverage other language models to extract conspiracy text or amplify their generation in natural language applications. However, we believe bringing light to the issue of conspiracy theory memorization in NLG models is essential for research to progress in the direction of safe and fair natural language processing and will enable future research to utilize these studies in model interpretability. 

While it may be possible for attackers to extract conspiracy theories from language models through more advanced techniques, we attempt to study how the ordinary user may fall prey to this type of information in a standard way. With language models becoming increasingly integrated into everyday natural language processing applications, the risks of unintentionally generating and spreading conspiracy theories rises. Our work can help researchers and engineers of language models thoroughly test these models for the generation of harmful conspiracy theory text using our analysis. We also hope to show researchers that even seemingly ``clean'' datasets may not diminish harmful noise from data and instead reflect or even amplify it after training.

\bibliography{acl2021}
\bibliographystyle{acl_natbib}

\appendix
\section{Appendix} \label{sec:appendix}
\subsection{Mechanical Turk Experiments}
Our conspiracy theory affirmation task was grouped into batches of 5 statements per topic and workers were compensated at 22¢ per batch. For the conspiracy topic task, we provided a topic, e.g. Holocaust, and ask the worker to click ``yes'' if they have heard of a conspiracy theory related to the topic. This simple task was paid at 2¢ per topic. The first task required workers to be located in the United States while the second task had no location restriction. For both tasks, the time it takes to read directions and answer the questions enabled workers to earn roughly \$9/hour. This is currently well above the average worker compensation on Mechanical Turk of \$2-5/hour \cite{10.1145/3173574.3174023,hitlin2016research} which states only 4\% of workers earn more than the U.S. federal minimum wage of \$7.25/hour. The second task, which contains no location restriction, may have workers from countries with a lower minimum wage. In Figures \ref{fig:ref} and \ref{fig:instructions} we provide screenshots of an example of our conspiracy theory affirm task, including the topic reference text and worker instructions.

\begin{figure}[h]
	\centering
    \begin{mdframed}
    {\large \textbf{Instructions}}\vspace{2pt}\\
    {\small Please Read the following reference text describing various conspiracy theories about a topic and answer the following questions.}\\
    {\textbf{Reference Text}}\vspace{2pt}\\
    {\small Moon landing conspiracy theories claim that some or all elements of the Apollo program and the associated Moon landings were hoaxes staged by NASA, possibly with the aid of other organizations. The most notable claim is that the six crewed landings (1969-1972) were faked and that twelve Apollo astronauts did not actually walk on the Moon. Various groups and individuals have made claims since the mid-1970s that NASA and others knowingly misled the public into believing the landings happened, by manufacturing, tampering with, or destroying evidence including photos, telemetry tapes, radio and TV transmissions, and Moon rock samples.}
    \end{mdframed}	\caption{Mechanical Turk conspiracy theory affirmation reference text.}
	\label{fig:ref}
\end{figure}

\begin{figure}[h]
	\centering
    \begin{mdframed}
    {\textbf{Questions}}\vspace{2pt}\\
    {\small \textbf{For the next five statements, evaluate whether or not they state/affirm any conspiracy theories in the reference text. If any part of the initial text states or affirms any ideas in the reference text, click yes. Also, if any part of the initial text follows the general nature of conspiracy theories in the reference text such as ``The moon landing is fake'', click yes. Otherwise, click no. Some statements may be the same but you should still evaluate each one.}}
    \end{mdframed}	\caption{Mechanical Turk conspiracy theory affirmation instructions.}
\label{fig:instructions}
\end{figure}

%\begin{figure*}[t]
%  \centering
%  \includegraphics[width=0.96\linewidth]{moon.png}
%  \caption{Mechanical Turk conspiracy theory affirmation reference text.}\label{fig:ref}
%\end{figure*}

%\begin{figure*}[t]
%  \centering
%  \includegraphics[width=0.96\linewidth]{instructions.png}
%  \caption{Mechanical Turk conspiracy theory affirmation instructions.}\label{fig:instructions}
%\end{figure*}

\end{document}